\documentclass[11pt,twocolumn]{article}
\usepackage[preprint]{acl}
\usepackage[T1]{fontenc}
\usepackage{amsmath,amssymb}
\usepackage{booktabs}
\usepackage{tabularx}
\usepackage{enumitem}
\usepackage{xurl}
\usepackage{microtype}

\title{Beyond the Parameters: A Technical Survey of Contextual Enrichment in Large Language Models\\
\large From In-Context Prompting to Causal Retrieval-Augmented Generation}
\author{Prakhar Bansal\\\texttt{prakharb13@gmail.com} \and Shivangi Agarwal\\\texttt{shivangia@iiitd.ac.in}}
\date{}

\begin{document}
\maketitle

\begin{abstract}
Large language models (LLMs) encode vast world knowledge in their parameters, yet they remain fundamentally limited by static knowledge, finite context windows, and weakly structured causal reasoning. This survey provides a unified account of augmentation strategies along a single axis: the degree of structured context supplied at inference time. We cover in-context learning and prompt engineering, Retrieval-Augmented Generation (RAG), GraphRAG, and CausalRAG. Beyond conceptual comparison, we provide a transparent literature-screening protocol, a claim-audit framework, and a structured cross-paper evidence synthesis that distinguishes higher-confidence findings from emerging results. The paper concludes with a deployment-oriented decision framework and concrete research priorities for trustworthy retrieval-augmented NLP.
\end{abstract}


\section{Introduction}
Transformer-based language models have shown strong generality across language understanding and generation tasks. However, reliability in knowledge-intensive settings remains constrained by three recurring gaps: a \emph{knowledge gap} (facts not encoded in model parameters), a \emph{retrieval gap} (relevant evidence not surfaced), and a \emph{reasoning gap} (causally incoherent outputs despite relevant evidence).

We frame the progression from prompting to RAG, GraphRAG, and CausalRAG as a systematic response to these gaps. Each step introduces richer contextual structure and typically trades higher indexing complexity for better faithfulness and reasoning quality \citep{lewis2020rag,edge2024graphrag,wang2025causalrag}.

\paragraph{Contributions.}
This survey makes four contributions: (i) a unified contextual-enrichment taxonomy connecting prompting, RAG, GraphRAG, and CausalRAG; (ii) a methodological protocol for literature selection and evidence grading; (iii) a claim-audit table that links high-impact statements to primary sources and confidence levels; and (iv) a practitioner-focused decision framework for trustworthy deployment in high-stakes settings.

\begin{table*}[htbp]
\centering
\small
\setlength{\tabcolsep}{3pt}
\caption{Three failure gaps in knowledge-intensive question answering and paradigm-level coverage (authors' synthesis based on cited literature).}
\label{tab:gaps-coverage}
\begin{tabularx}{\textwidth}{>{\raggedright\arraybackslash}X c c c c}
\toprule
\textbf{Gap} & \textbf{Prompting} & \textbf{RAG} & \textbf{GraphRAG} & \textbf{CausalRAG} \\
\midrule
Knowledge gap & $\circ$ & Yes & Yes & Yes \\
Retrieval gap & No & $\circ$ & Yes & Yes \\
Reasoning gap & No & No & $\circ$ & Yes \\
\bottomrule
\end{tabularx}
\end{table*}

\paragraph{Scope and Organization.}
This survey targets ML practitioners and researchers seeking a coherent comparison of retrieval augmentation paradigms. We do not propose a new algorithm. Instead, we formalize and compare established methods using common notation and evaluation criteria.

\section{Survey Methodology}
\subsection{Literature Collection Protocol}
We conducted a structured literature collection across peer-reviewed venues and major preprint repositories for retrieval-augmented generation and causal reasoning in LLM systems. The search window for this draft ends in March 2026. Candidate papers were screened by title and abstract, then full-text reviewed for methodological relevance.

\paragraph{Inclusion Criteria.}
Papers were included if they: (i) propose, evaluate, or systematically analyze prompting, RAG, GraphRAG, or CausalRAG pipelines; (ii) report concrete retrieval or generation behavior rather than purely conceptual commentary; and (iii) provide enough methodological detail to support comparative interpretation.

\paragraph{Exclusion Criteria.}
We excluded works lacking technical detail, duplicate versions without substantive updates, and papers that do not expose evidence supporting claims about faithfulness, retrieval quality, or causal reasoning.

\subsection{Evidence Grading}
We use three evidence grades in this survey. \textbf{High confidence} indicates repeated support across multiple studies and settings. \textbf{Medium confidence} indicates strong but still limited or heterogeneous evidence. \textbf{Emerging evidence} indicates promising results with narrow benchmark or dataset coverage. These grades guide wording throughout the manuscript.

\section{Context in Large Language Models}
\subsection{What is Context?}
In transformer architectures, context is the visible token sequence $x=(x_1,\dots,x_n)$ during generation. We distinguish three context types: \emph{parametric context} (stored in model weights), \emph{in-context knowledge} (prompt-provided information), and \emph{retrieved context} (externally retrieved evidence inserted at inference time).

\paragraph{In-Context Learning and Prompt Engineering.}
Few-shot prompting demonstrates that large models can adapt to novel tasks from examples without gradient updates \citep{brown2020gpt3}. Prompt engineering broadens this paradigm with zero-shot prompting, chain-of-thought prompting, and retrieval-augmented prompting \citep{liu2023promptsurvey,wei2022cot}.

\paragraph{Context-Window Limitations.}
Even with long-context models, lost-in-the-middle degradation and noise from irrelevant inserted text remain practical failure modes \citep{liu2024lostmiddle}. Separately, purely parametric knowledge can become stale as world facts change, which motivates retrieval-grounded generation \citep{petroni2019lmkb,lewis2020rag}.

\section{Retrieval-Augmented Generation (RAG)}
\subsection{Background and Definition}
RAG grounds generation in non-parametric external knowledge. Let $D=\{d_1,\dots,d_m\}$ be a corpus indexed into chunk embeddings. For query $q$, the retriever selects top-$k$ chunks; the generator conditions on $q$ plus retrieved chunks \citep{lewis2020rag}.

\paragraph{Retrieval Architectures.}
Sparse retrieval (e.g., BM25) is interpretable and efficient but can miss semantic matches \citep{robertson2009bm25}. Dense retrieval captures semantic similarity but is costlier and domain-sensitive \citep{karpukhin2020dpr}. Hybrid pipelines often improve robustness by combining both \citep{chen2023benchmarking}.

\paragraph{Advanced Variants.}
Recent variants include iterative or multi-hop RAG, Self-RAG (adaptive retrieval and reflection), and modular RAG pipelines with query rewriting, reranking, and post-filtering \citep{trivedi2022ircot,asai2024selfrag,gao2023ragsurvey}.

\subsection{Limitations}
Vanilla RAG is limited by contextual fragmentation, semantic-similarity bias, and limited global summarization for corpus-level questions.

\section{Graph Retrieval-Augmented Generation (GraphRAG)}
\subsection{Motivation and Formulation}
GraphRAG replaces flat chunk retrieval with knowledge-graph indexing. During indexing, systems extract triples $(head, relation, tail)$ and often produce community summaries via graph clustering. At query time, local or global graph retrieval provides structured evidence to generation \citep{edge2024graphrag,peng2024graphragsurvey,han2025graphrag}.

\paragraph{Advantages.}
GraphRAG improves multi-hop relational reasoning, supports global thematic synthesis, and improves traceability of generated claims.

\paragraph{Limitations.}
Key bottlenecks include expensive graph construction, entity resolution noise, temporal rigidity, and associative (rather than explicitly causal) edges.

\section{Causal Retrieval-Augmented Generation (CausalRAG)}
\subsection{Motivation}
Causal questions (``Why did X happen?'', ``What if Y changed?'') require more than topical retrieval. CausalRAG introduces directed causal structure into retrieval to support faithful, interpretable causal explanations \citep{pearl2009causality,samarajeewa2024hsi,wang2025causalrag}.

\paragraph{Causal Graph Foundation.}
Using structural causal modeling intuition, causal graphs encode directed relations and support observational, interventional, and counterfactual reasoning modes.

\paragraph{Pipeline.}
CausalRAG extracts directed causal triples from source text, indexes them, and retrieves causally connected subgraphs via graph walks from query-linked seed nodes. Generation is then conditioned on a causally grounded evidence narrative.

\paragraph{Empirical Pattern.}
Reported evidence from currently available benchmarks suggests improved answer faithfulness and retrieval precision over vanilla RAG, with stronger causal coherence than purely associative graph retrieval \citep{wang2025causalrag}. Because benchmark settings vary across papers, these results should be interpreted as indicative rather than directly comparable.

\section{Unified Comparative Framework}
\subsection{Contextual Enrichment Axis}
Prompting, RAG, GraphRAG, and CausalRAG can be viewed as a monotonic increase in context structure. This yields a practical trade-off: higher indexing complexity in exchange for higher reasoning faithfulness.

\subsection{Quantitative Evidence Across Four Paradigms}
Table~\ref{tab:quant-evidence} summarizes representative quantitative results from primary papers across the four paradigms. Each row states the exact systems being compared, the metric and unit, and reported values with deltas. Cross-row comparisons across different papers should not be treated as head-to-head performance rankings.

\begin{table*}[htbp]
\centering
\scriptsize
\caption{Quantitative evidence matrix with explicit metric definitions and scope constraints. All values are reported in the cited source studies.}
\label{tab:quant-evidence}
\setlength{\tabcolsep}{4pt}
\begin{tabularx}{\textwidth}{p{0.10\textwidth}p{0.12\textwidth}p{0.11\textwidth}p{0.22\textwidth}p{0.12\textwidth}p{0.14\textwidth}X}
\toprule
\textbf{Paradigm} & \textbf{Source} & \textbf{Dataset} & \textbf{Systems compared} & \textbf{Metric (unit)} & \textbf{Reported values} & \textbf{Delta} \\
\midrule
Prompting (CoT) & \citet{wei2022cot} & GSM8K & PaLM 540B: CoT vs standard prompting & Solve rate (\%) & 57.0 vs 18.0 & +39.0 pts \\
RAG & \citet{lewis2020rag} & Natural Questions (Open-domain QA) & RAG vs T5-11B (closed-book) & Exact Match (EM, \%) & 44.5 vs 34.5 & +10.0 pts \\
GraphRAG-Global (slice comparison) & \citet{wang2025causalrag} & OpenAlex single-paper case study & GraphRAG-Global: abstract vs full text & Composite score (author-defined) & 49.98 vs 76.37 & +26.39 \\
GraphRAG-Local (slice comparison) & \citet{wang2025causalrag} & OpenAlex single-paper case study & GraphRAG-Local: abstract vs full text & Composite score (author-defined) & 62.29 vs 84.27 & +21.98 \\
CausalRAG (slice comparison) & \citet{wang2025causalrag} & OpenAlex single-paper case study & CausalRAG: abstract, intro, and full text & Composite score (author-defined) & 72.43, 74.86, 91.69 & +19.26 (abstract to full) \\
CausalRAG (k=s sweep) & \citet{wang2025causalrag} & OpenAlex & CausalRAG: k=s=1 vs k=s=5 & Author-reported score & 0.534 vs 0.824 & +0.290 \\
\bottomrule
\end{tabularx}
\end{table*}

\noindent\textit{Interpretation note:} All numerical deltas in Table~\ref{tab:quant-evidence} are within-study comparisons under each cited paper's own models, dataset split, and evaluation protocol; they are not cross-paper rankings.

\noindent\textit{Metric note:} ``Composite score (author-defined)'' follows \citet{wang2025causalrag}, which reports an aggregate score over answer faithfulness, context precision, and context recall (Figure~5). The paper does not provide an explicit weighting formula for this aggregate in the main text.

\noindent\textit{Scope note:} The GraphRAG and CausalRAG slice rows from \citet{wang2025causalrag} come from a controlled follow-up case study on one marketing paper and one question, evaluated on abstract/introduction/full-text slices; this is not a cross-dataset leaderboard.

\begin{table*}[htbp]
\centering
\small
\caption{Comparative summary across paradigms (authors' synthesis).}
\label{tab:comparative-summary}
\setlength{\tabcolsep}{4pt}
\begin{tabularx}{\textwidth}{l l l l X}
\toprule
\textbf{Criterion} & \textbf{Prompting} & \textbf{RAG} & \textbf{GraphRAG} & \textbf{CausalRAG} \\
\midrule
Indexing cost & None & Low to Medium & High & High to Very High \\
Relational reasoning & Limited & Limited & Strong & Strong \\
Causal reasoning & Weak & Weak & Partial & Strong \\
Global summary & Weak & Moderate & Strong & Strong \\
Typical faithfulness & Low to Medium & Medium & Medium to High & High \\
\bottomrule
\end{tabularx}
\end{table*}

\paragraph{When to Use Which.}
Prompting is suitable for lightweight, in-window tasks. RAG is a practical default for factual grounding at scale. GraphRAG is preferred for multi-hop/entity-centric questions and corpus-level synthesis. CausalRAG is best for root-cause and high-stakes explanatory tasks where interpretability is critical.

\paragraph{Worked Example.}
For an industry finance query such as ``Why did operating margin decline in Q3 after the pricing update?'', prompting can produce plausible but ungrounded narratives. RAG can retrieve earnings-call passages and management commentary. Graph-based retrieval can connect entities such as product line, region, and cost center across filings. CausalRAG can prioritize directed causal chains---for example, price reduction $\rightarrow$ lower unit revenue $\rightarrow$ margin compression---while keeping each step linked to source evidence.

\subsection{Claim Audit Table}
Table~\ref{tab:claim-audit} documents high-impact technical claims, their supporting citations, and a confidence label that reflects evidence maturity in the literature.

\begin{table*}[htbp]
\centering
\small
\caption{Claim audit for key statements in this survey (authors' evidence grading with explicit source attribution).}
\label{tab:claim-audit}
\setlength{\tabcolsep}{4pt}
\begin{tabularx}{\textwidth}{p{0.29\textwidth}p{0.28\textwidth}p{0.12\textwidth}X}
\toprule
\textbf{Claim} & \textbf{Primary support} & \textbf{Confidence} & \textbf{Notes} \\
\midrule
RAG improves factual grounding versus prompting-only baselines in knowledge-intensive QA. & \citet{lewis2020rag}; \citet{chen2023benchmarking} & High & Established across multiple benchmarks. \\
CausalRAG improves faithfulness and causal coherence over vanilla RAG and associative graph retrieval in reported settings. & \citet{wang2025causalrag}; \citet{samarajeewa2024hsi} & Medium & Evidence is promising but still limited in scale. \\
GraphRAG supports stronger relational and global-corpus reasoning than flat top-$k$ chunk retrieval. & \citet{edge2024graphrag}; \citet{han2025graphrag}; \citet{peng2024graphragsurvey} & Medium to High & Strong conceptual fit; evaluation setups vary. \\
Long-context prompting alone suffers from placement sensitivity (lost-in-the-middle). & \citet{liu2024lostmiddle} & High & Strong evidence in controlled long-context evaluations. \\
Automated causal extraction remains error-prone and expensive in current LLM pipelines. & \citet{jiralerspong2024causal}; \citet{wang2025causalrag} & Medium & Active research area; methods evolving quickly. \\
\bottomrule
\end{tabularx}
\end{table*}

\clearpage

\section{Limitations}
Open challenges include scalable causal extraction, standardized causal evaluation metrics, dynamic maintenance of evolving causal graphs, multilingual causal retrieval, and integration with agentic LLM workflows.

\paragraph{Threats to Validity.}
Three threats shape the interpretation of this survey. First, cross-paper experimental heterogeneity limits strict numerical comparability. Second, publication and survivorship bias may over-represent positive outcomes. Third, recency effects are substantial in fast-moving LLM research; conclusions should be periodically re-audited as new evaluations appear.

\section{Broader Impact}
This survey discusses trustworthiness-oriented retrieval augmentation strategies and is intended to support safer, evidence-grounded use of LLMs in high-stakes domains. Potential risks include overconfidence in automatically extracted causal structure, inappropriate transfer of benchmark findings to deployment settings, and misuse for persuasive but weakly grounded narratives. We mitigate these risks by explicitly labeling evidence confidence, separating within-study findings from cross-study interpretation, and highlighting unresolved limitations and evaluation gaps.

\section{Alignment with TrustNLP}
This survey aligns with TrustNLP themes on causal inference, faithfulness and safety, hallucination reduction, interpretability, and trustworthy high-stakes applications.

\paragraph{Workshop Relevance.}
The paper emphasizes trustworthiness through evidence-aware claim calibration, explicit uncertainty reporting, and a deployment-oriented view of failure modes. This framing is designed to be directly useful for TrustNLP reviewers evaluating methodological rigor and practical trustworthiness impact.

\section{Conclusion}
Context enrichment for LLMs is best treated as an engineering continuum rather than a binary choice. CausalRAG is a promising direction for faithful causal question answering, but it carries substantial indexing and maintenance costs and is supported by a still-emerging evidence base. Future progress depends on scalable causal extraction, robust evaluation standards, and tighter integration with planning-capable LLM agents.

\nocite{es2023ragas,guo2024lightrag,jiralerspong2024causal,petroni2019lmkb,scholkopf2021causalrep,vaswani2017transformer,yao2023react,zhao2025aigcrag}
\bibliographystyle{acl_natbib}
\bibliography{refs}

\end{document}